# Relational reasoning networks

Giuseppe Marra [a], Michelangelo Diligenti [b,*], Francesco Giannini [c]

[a] Department of Computer Science, KU Leuven, Belgium
[b] Department of Information Engineering and Mathematics, University of Siena, Italy
[c] Faculty of Sciences, Scuola Normale Superiore, Italy

ABSTRACT

Neural-symbolic methods integrate neural architectures, knowledge representation and reasoning. However, they have struggled with both the intrinsic uncertainty of the observations and scaling to real-world applications. This paper presents Relational Reasoning Networks (R2N), a novel end-to-end model that performs relational reasoning in the latent space of a deep learner architecture, where the representations of constants, ground atoms and their manipulations are learned in an integrated fashion. Unlike flat architectures such as Knowledge Graph Embedders, which can only represent relations between entities, R2Ns define an additional computational structure, accounting for higher-level relations among the ground atoms. The considered relations can be explicitly known, like the ones defined by logic formulas, or defined as unconstrained correlations among groups of ground atoms. R2Ns can be applied to purely symbolic tasks or as a neural-symbolic platform to integrate learning and reasoning in heterogeneous problems with entities represented both symbolically and feature-based. The proposed model overtakes the limitations of previous neural-symbolic methods that have been either limited in terms of scalability or expressivity. The proposed methodology is shown to achieve state-of-the-art results in different experimental settings.

## 1. Introduction

Enhancing deep learning models with explicit reasoning abilities represents a key challenge in the AI field, essential for achieving semantically accurate, robust, and reliable AI systems [1,2]. In particular, Neural-Symbolic (NeSy) AI [3–6] combines the strengths of neural models, which are highly effective in exploiting complex latent patterns over large collections of data, and symbolic reasoning, which excels at representing and performing inference on structured knowledge. By integrating these two paradigms, NeSy methods can leverage the capabilities of deep learning architectures to automatically extract meaningful feature representations, while also incorporating symbolic reasoning mechanisms for handling structured knowledge.

NeSy systems are often applied over relational data, like Knowledge Graphs (KG), where structured representations express complex relationships and dependencies among different entities, facilitating a more nuanced understanding and reasoning within the domain. Among the existing approaches for relational domains, a powerful class of methodologies [7–9] uses logic formulas as templates for Probabilistic Graphical Models (PGM), which explicitly encode statistical dependencies among entities and their relationships. However, the application of these methods has been limited to small scale tasks. In fact, application to large-scale domains is hindered by the high computational complexity of performing exact probabilistic inference on the corresponding probabilistic model.

On the other hand, Knowledge Graph Embeddings (KGE) [10–12] provide a scalable solution to reasoning over knowledge graphs. KGE methods first embed the entities and relations, and then classical distance-based functions are used to generalize to the missing facts. KGEs focus on modeling statistical regularities among relations and entities but their strong assumptions make them fail at detecting and exploiting higher level logical knowledge, like formulas or programs.

Different NeSy models have been recently proposed to combine knowledge graph reasoning with logic rules and neural architectures. In particular, some approaches inject the logic knowledge into the learning objective on top of a KGE model [13,14], or use the logic rules externally to augment the labels for training the KGE [15]. Another line of research uses a neural architecture to first learn a set of relevant logic rules, e.g. via a recurrent neural network from the known KG [16], or to assign a score to each possible relation in a given relational path [17]. However, none of these systems allows for a unified integration between the representations obtained with neural architectures and knowledge graph embeddings, which can also take into account the injection of a set of logic rules. Overall, the definition of a method capable of exploiting logic knowledge in a relational domain, while still being able to scale to large domains, is still an open challenge for the NeSy research community.

* Corresponding author.
*E-mail addresses:* giuseppe.marra@kuleuven.be (G. Marra), michelangelo.diligenti@unisi.it (M. Diligenti), francesco.giannini@sns.it (F. Giannini).

This paper introduces Relational Reasoning Networks (R2N), a class of models that produces semantically refined embedding representations, obtained by exploiting logic knowledge in a relational domain. The presented work represents a significant step towards devising a model that can effectively and efficiently preserve and integrate the benefits of embedding representations, probabilistic logic inference and neural architectures. In particular, R2Ns jointly develop integrated representations for constants, ground atoms and ground knowledge by computing multiple sub-symbolic reasoning steps in a latent space. The learning of embedding representations allows the model to automatically select whether and how much some specific logic knowledge is useful for the specific task. As a result, R2Ns maintain enough expressive power to perform relational reasoning under uncertainty, while allowing to scale up to larger relational settings compared to PGM-based approaches. R2N inference mechanism is divided into three phases. Firstly, symbolic ground atoms are embedded in a latent space using representation learning techniques, e.g. KGE. Secondly, the relations among atoms provided in the logic theory are used to sub-symbolically aggregate and refine the atom embeddings in a multi-layer fashion. This phase resembles multi-hop automated reasoning, but in a latent space. Lastly, the final embedding representations are used to make predictions. These predictions can be used to train the model end-to-end with respect to a supervised task.

**Contributions.** The main contribution of the paper is the introduction of a neural-symbolic architecture that performs relational reasoning in a latent space. Distilling the relational structure at the embedding level allows us to perform both learning and reasoning in a more scalable way. The experimental results show the effectiveness of the approach in achieving state-of-the-art results in different experimental settings. In addition, the proposed methodology is very general, as it can be used both in relational tasks where the prior logic knowledge is explicitly available, even if noisy (e.g. predictions can deviate from the knowledge with a variable degree that must be co-learned), and in tasks where the relational structure is present but left latent. Finally, the methodology provides a general platform for neural-symbolic integration [6,18], as it can be transparently applied to pure symbolic inputs, or to cases where the inputs have a feature-based representation, like images.

The outline of the paper is the following: Section 2 summarizes basic notions relevant to define our model, while Section 3 explicitly formulates the problem statement that is solved by the proposed model. Section 4 presents the model and how it approximates the reasoning process, Section 5 shows the experimental results on multiple datasets and learning tasks, and Section 6 provides insights and theoretical guarantees for the reasoning process implemented by R2Ns. Finally, Section 7 presents an overview of related works, while some conclusions and remarks on future works are drawn in Section 8.

## 2. Preliminary

### 2.1. First-order logic

We consider a function-free First-Order Logic (FOL) language $\mathcal{L}$, composed of a finite set of constants $C$ for specific domain entities, a set $\mathcal{X}$ of variables for anonymous entities and a set $\mathcal{P}$ of $n$-ary predicates for relations among constants. Given an $n$-ary predicate $P$ and a tuple $(t_1, \ldots, t_n)$ of constants or variables, $P(t_1, \ldots, t_n)$ is called an *atom*. If $t_1, \ldots, t_n$ are only constants, $P(t_1, \ldots, t_n)$ is called a *ground atom*. The set of all the possible ground atoms of a FOL language is called Herbrand Base (HB). Logic rules are defined as usual by using logic connectives $\{\neg, \wedge, \vee, \rightarrow\}$ and quantifiers $\{\forall, \exists\}$.

**Example 1** (*FOL Language*). Consider a set of constants $C = \{a, b, c\}$ representing people (e.g. (a)lice, (b)ob and (c)arl), and the predicate set $\mathcal{P} = \{S(\cdot), F(\cdot,\cdot)\}$ stating if a person (S)mokes or if two people are (F)riends, respectively. The Herbrand Base of this language is composed of 12 ground atoms, obtained by grounding the predicate $S$ over the constants in $C$ and the predicate $F$ over the pair of constants in $C$, i.e. $HB = \{S(a), S(b), S(c), F(a, a), F(a, b), \ldots, F(c, c)\}$.

Our approach considers a logic theory $\mathcal{T}$ as a set of unquantified FOL rules $r_i$, each depending on a set of variables $X_i \subset \mathcal{X}$, i.e. $\mathcal{T} = \{r_1(X_1), r_2(X_2), \ldots, r_m(X_m)\}$. A substitution $\theta$ for $X_i$ is a replacement of constants in $C$ to the variables in $X_i$. Given a rule $r_i(X_i)$ and a substitution $\theta$ for $X_i$, we obtain a ground rule by applying the substitution $\theta$ to the variables in $X_i$. *Grounding* a rule $r_i(X_i)$ refers to the process of applying all the possible substitutions $\Theta$ for $X_i$ given $C$, where $|\Theta| = |C|^{|X_i|}$. We indicate by $R_i$ the set of all the groundings of $r_i(X_i)$. Grounding the entire $\mathcal{T}$ refers to grounding all its rules and the resulting set is indicated by $R$.

**Example 2** (*Logic Theory*). Consider $\mathcal{T} = \{r_1(X_1)\}$, with $r_1(X_1) = S(x) \wedge F(x, y) \rightarrow S(y)$ stating that "if a person $x$ smokes and $x$ and $y$ are friends, then also $y$ smokes." Here, $X_1 = \{x, y\}$ is the set of variables of $r_1$. For instance, we can ground $r_1(X_1)$ by the substitution $\theta = \{x : a,\ y : b\}$ for $X_1$, thus obtaining $S(a) \wedge F(a, b) \rightarrow S(b)$. If $C = \{a, b, c\}$ we get $R_1 = \{S(a) \wedge F(a, a) \rightarrow S(a), S(a) \wedge F(a, b) \rightarrow S(b), \ldots, S(c) \wedge F(c, c) \rightarrow S(c)\}$, with $|R_1| = 3^2 = 9$.

Logic theories can benefit from graph representations to define inference and learning algorithms according to their underlying relational structures. In this paper, we encode two components of our logic language as graphs: (i) the set HB of ground atoms as an n-ary knowledge graph (Section 2.2); (ii) the set of ground rules $R$ as a factor graph (Section 2.3).

### 2.2. Knowledge graph embeddings

Knowledge graphs (KG) are graph data structures representing relational knowledge consisting of facts (ground atoms) in form of triples formed by two entities (constants) and a relation (binary predicate). In a KG each considered entity is a node and each fact is a relation establishing an edge between two entities. KGs are incomplete and Knowledge Graph Embeddings (KGE) are a powerful approach for populating KGs by mapping entities and relations to a latent representation, which generalizes the assignments to unknown facts. Indeed, KGE methods learn the entity and relation embeddings by defining scoring functions that are trained to match the supervisions.

**Example 3** (*KGE*). Let $F(a, b)$ be a fact of a KG with $e_a, e_b, W_F$ indicating trainable embedding vectors for entities $a, b$ and the relation $F$, respectively. *TransE* [19] is a well-known KGE tha models relations as translation operations on the embeddings of the entities, where the internal atom representation $e_a + W_F - e_b$ is scored as $1/(1 + \|e_a + W_F - e_b\|)$.

KGs can be extended to an HB with generic $n$-ary relations, where the bipartite graph representation is generalized such that an $n$-ary atom establishes $n$ edges from the atom node to the $n$ entities appearing in the atom, see e.g. Fig. 1-(a). Many KGE approaches can also be trivially extended to $n$-ary relations as discussed by Fatemi et al. [20].

### 2.3. Encoding grounded FOL formulas as factor graphs

A common practice in the field of Statistical Relational AI (StarAI) [21] is to map FOL theories to undirected probabilistic graphical models (Fig. 2-a), represented via factor graphs [22] (Fig. 2-b), to get advantage of the available standard tools for model training and inference. This paper exploits the same translation used by Markov Logic

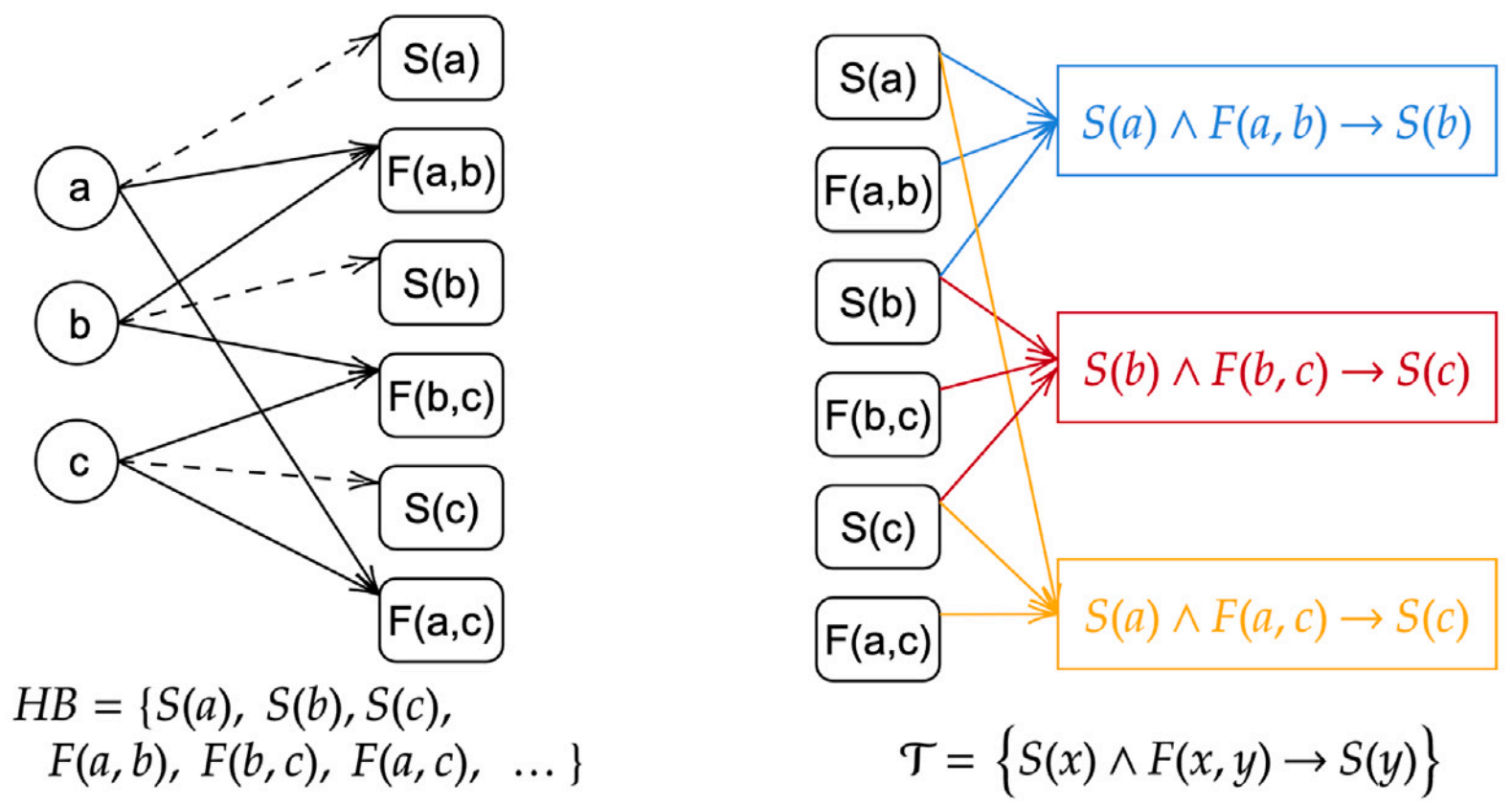

(a) $n$-ary KGE                    (b) Logic-based Factor graph

**Fig. 1.** **(a)** An $n$-ary KGE where different arrows correspond to different predicates, here we used the dashed one for $S$ and the solid one for $F$. **(b)** A factor graph for the logic theory $\mathcal{T} = \{S(x) \wedge F(x,y) \rightarrow S(y)\}$, grounded for the constants $\{a,b,c\}$, where each ground formula is highlighted with a different color. In both graph representations some nodes/edges have been omitted for readability reasons.

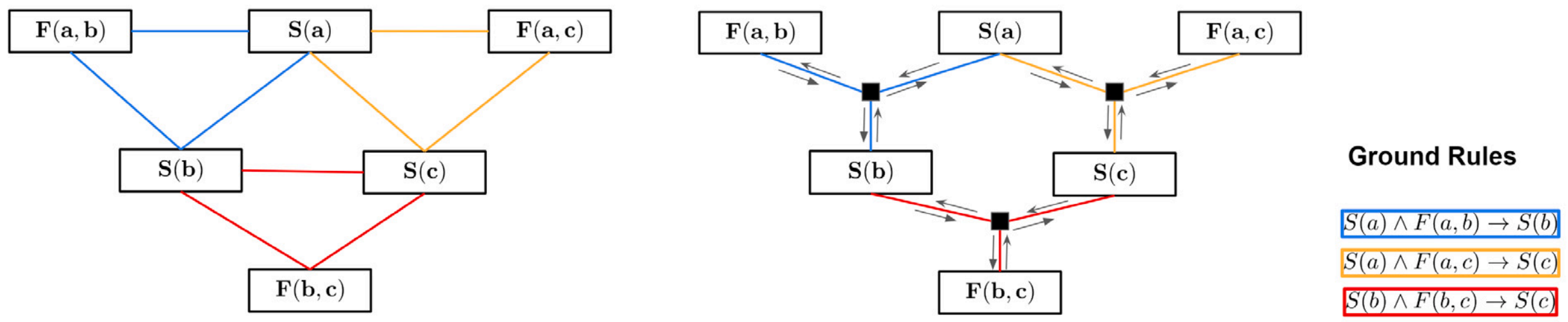

a) Undirected graphical model                    b) Factor graph

**Fig. 2.** (a) The portion of undirected graphical model corresponding to the logic rule $\forall x \forall y S(x) \wedge F(x,y) \rightarrow S(y)$ on three different groundings. b) The factor graph corresponding to the graphical model in a), factor nodes correspond to the black squares, each associated to a clique representing a ground formula.

Networks (MLN) [23], mapping a logic theory into a factor graph by grounding all the rules.

A factor graph $(V, F, \mathcal{E})$ is a bipartite undirected graph, whose vertices are divided into two disjoint sets: the variable nodes $V$ and the factor nodes $F$. Edges in $\mathcal{E}$ connect nodes in $V$ to nodes in $F$. A *logic-based factor graph* can be built given the HB of a FOL language and a logic theory $\mathcal{T}$ by:

- adding a variable node $a$ for each ground atom: $V = HB$;
- adding a factor node $g$ for each ground rule: $F = R$;
- adding an edge $(a, g)$ in $\mathcal{E}$ if the atom associated to $a$ occurs in the ground rule associated to $g$.

Given an atom $a$ and a ground rule $g$, $ne(a)$ and $ne(g)$ indicate the list of neighbor nodes of $a$ and $g$, respectively. $ne(a,i)$ and $ne(g,i)$ denote the $i$th elements of such lists, respectively.

**Example 4** (*Logic-based Factor Graph*)**.** A portion of the logic-based factor graph corresponding to the FOL theory $\mathcal{T}$ in Example 2 is represented in Fig. 1-(b).

Efficient message passing schema can be defined over the factor graph, like for example the Belief Propagation algorithm [24]. Fig. 2-b shows the factor graph corresponding to the undirected graph in Fig. 2-a, showing how the inference can be decomposed as a message passing schema from the variable nodes to the factor nodes and back to the variables. Section 4 describes the inference process of R2Ns by adopting a neural parameterization of message passing, where the unfolding of the logic-based factor graph corresponds to the architecture of the reasoning layers within the R2N (Section 4.2).

## 3. The problem

**Problem Definition. Given:**

- a FOL language $\mathcal{L} = (\mathcal{C}, \mathcal{P})$ and a logic theory $\mathcal{T}$;
- the ground truth $y_e$ for $E \subset$ HB (i.e. the evidence)
- (optional) a feature representation $I$ of constants in $\mathcal{C}$;

**Find:**

- the truth values $y_Q$ for $Q \subset$ HB (i.e. the queries).

Examples of instances of the problem:

- *Classification:* $\mathcal{C}$ is the set of input patterns and $I$ their feature representation, $\mathcal{P}$ is the set of classes, $\mathcal{T} = \emptyset$. $y_E$ are the class labels for the subset of the patterns in $\mathcal{C}$ (i.e. the training set). $y_Q$ are the class predictions for another subset of the patterns in $\mathcal{C}$ (i.e. the test set).color
- *Link Prediction:* $I = \mathcal{T} = \emptyset$. $E$ is the set of the known links on the graph, $Q$ is the set of unknown links to predict.
- *Logical Reasoning:* $I = \emptyset$, $\mathcal{T}$ is the logic theory. $E$ is the set of known logic facts. $Q$ is a subset of the conclusions of the reasoning process.
- *Neural-Symbolic:* all the elements of the previous points can be integrated to add logical reasoning or link prediction on top of a pattern recognition task.

## 4. The model

Relational Reasoning Networks (R2Ns) are neural-based models, whose internal inference structure mimic the structure of belief prop-agation over factor graphs. In particular, R2Ns are neural-symbolic reasoning models composed of three high-level components.

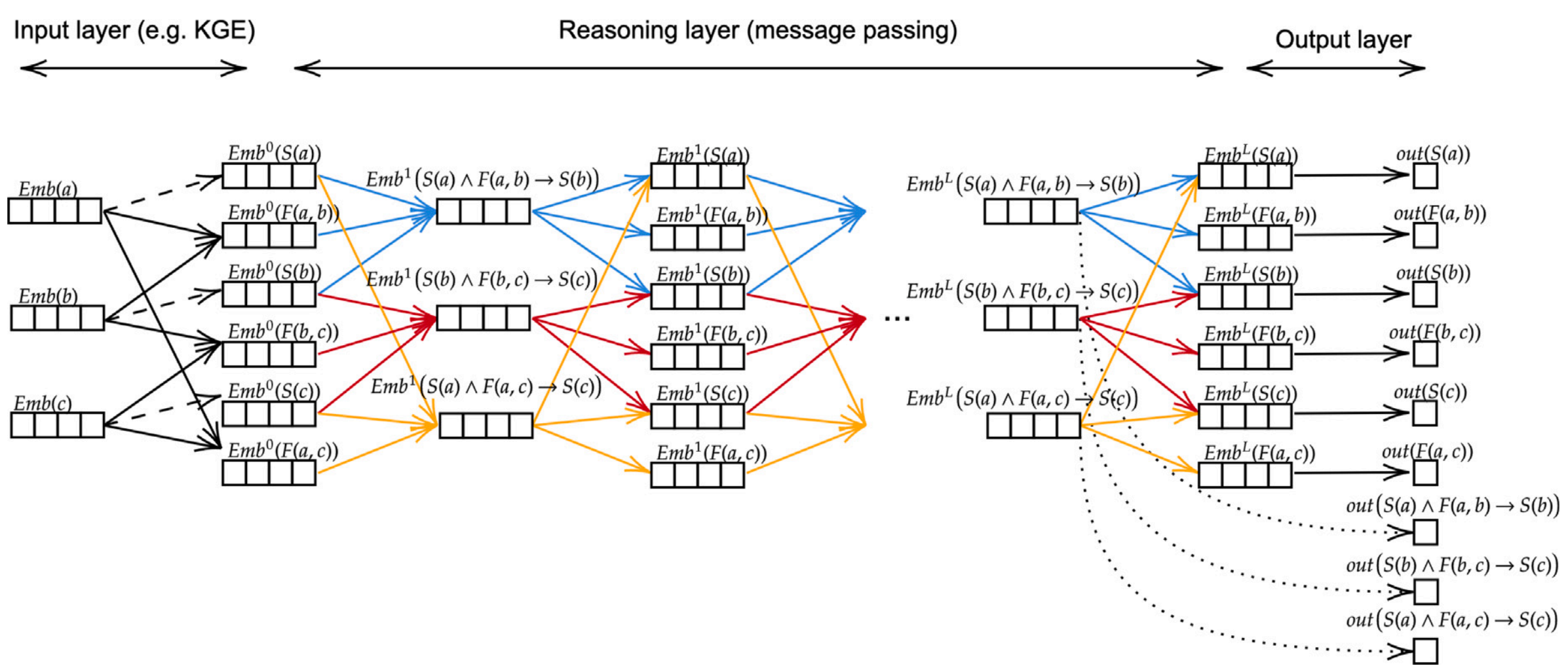

**Fig. 3.** Overall structure of the model. The model can be seamlessly used to: *(i) classify* whether $a \in C$ smokes by computing $out(S(a))$; *(ii) predict the link* stating whether $a, b \in C$ are friends by computing $out(F(a, b))$; and *(iii) reason* about $b \in C$ smoking because being a friend of the smoker $a$ by exploiting the rule $F(a, b) \wedge S(a) \rightarrow S(b)$.

1. The **Input Atom Embedding Layer** maps each ground atom $a_i \in$ HB into an embedding vector.
2. **Reasoning Layers** (recursively) update the atom embeddings according to the structure imposed by the logic-based factor graph associated to $\mathcal{T}$.
3. **Output Layers** use the final atom embeddings to take decisions on the query atoms in $Q$.

The overall model is shown in Fig. 3. Algorithm 1 details the step-by-step R2N computation, which is detailed in the next sections.

### *4.1. Input atom embedding layer*

The first layer of an R2N assigns a numeric representation to all the ground atoms in the HB. In particular, each ground atom is assigned an embedding vector in some *latent* space whose semantics is not defined a-priori. Given a ground atom $a = P(c_1, \dots, c_n)$ and $Emb(c_i)$ the embedding of $c_i \in C$, the input atom embedding layer computes the embedding $Emb^0(a)$ as:

$$Emb^0(a) = f_P(Emb(c_1), \dots, Emb(c_n)) \tag{1}$$

A vast range of representation learning approaches can be employed to implement and co-train the function $f_P$. In particular, in this paper we consider Knowledge Graph Embeddings and latent representations of deep neural networks.

*Knowledge graph embeddings.* KGEs (Section 2.2) encode entities and relations with real-valued embedding vectors in order to represent the relational structure among different objects in a meaningful latent space. All KGE methods internally perform two operations: reconstruct an atom representation from the representations of the constants and the predicate, and compute a score from this representation via a scoring function.

*Latent representations of deep networks.* In case a feature representation $I$ of the constants (cf. Section 3) is provided, such as with images, time series, videos, text, etc., we can obtain the atom embedding representations by direct extraction form hidden representations of neural architectures.

### *4.2. Reasoning layers*

Reasoning layers recursively take a HB representation as input and provide an updated HB representation as output. Inspired by recent advances in the Graph Neural Network (GNN) community [25], we implement reasoning layers as the unfolding network of a message passing process [26] over the logic-based factor graph. The message from an atom node $a$ to a ground rule node $g$ at layer $\ell$ is defined as:

$$M^{\ell}_{a \to g} = Emb^{\ell-1}(a) \tag{2}$$

where $Emb^{\ell-1}(a)$ is the embedding of the atom $a$ at layer $\ell - 1$, which resolves to the output of the input atom embedding layer for the first reasoning layer, i.e. for $\ell = 1$.

The embedding of the ground rule node $g \in R_j$ for rule $r_j$ at layer $\ell$ is then computed as a Multi-Layer Perceptron ($MLP$) on the messages of its neighbor atom nodes $ne(g)$:

$$Emb^{\ell}(g) = MLP^1_j(M^{\ell}_{ne(g,1) \to g}, \dots, M^{\ell}_{ne(g,d_j) \to g}) \tag{3}$$

where $d_j = |ne(g)|$, i.e. the number of atoms occurring in $r_j$. Notice that we have a different function $MLP^1_j$ for each rule $r_j$. This is important to enable the different rules to differently aggregate the incoming messages from their neighbor atoms.

The message from a ground rule node $g \in R_j$ to its $i$th neighbor atom node $a = ne(g, i)$ is defined as:

$$M^{\ell}_{g \to a} = MLP^2_{j,i}(Emb^{\ell}(g)) \tag{4}$$

where, the message function $MLP^2_{j,i}$ is different for each rule $r_j$ and for each position $i$ of the recipient node of the message.

Finally, the embedding of an atom node $a$ at layer $\ell$ is updated by aggregating the messages of all the ground rule nodes $g$, where $a$ is a neighbor of $g$:

$$Emb^{\ell}(a) = \underset{(g,i):\ a=ne(g,i)}{\mathcal{A}} M^{\ell}_{g \to a} \tag{5}$$

where $\mathcal{A}$ is an aggregation operator like sum, mean or max. In the experiments we used a summation aggregator.

The message passing Eqs. (2) to (5) implements an update schema for the atom embeddings, which can be repeated multiple times in a recursive fashion. Each atom embedding is initialized by the input atom embedding layer (see Section 4.1) and then updated multiple times by the message passing schema. Our message passing process is also inspired by standard logic reasoning, e.g. forward chaining, where the truth degree of the atoms is iteratively updated from the truth degree of other atoms appearing in the same rules.

**Algorithm 1:** The basic schema of the computation performed by an R2N.

**Input:**
$\mathcal{C}$: set of constants, $\mathcal{P}$: set of predicates, $\mathcal{T}$: logic theory, $L \geq 1$: number of reasoning layers

**Initialization**
$HB = \mathtt{HB}(\mathcal{C}, \mathcal{P})$ *// Build the Herbrand base (Section 2.1)*
$R = \mathtt{Ground}(\mathcal{T}, HB)$ *// Build the set of ground rules (Section 2.1)*

**Input Atom Embedding layer**
**for** $a = P(c_1, \ldots, c_n) \in HB$ **do**
  $Emb^0(a) = f_P(Emb(c_1), \ldots, Emb(c_n))$ *// Initialize atom emb. (Equation (1))*

**Reasoning layers**
**for** $l = 1, \ldots, L$ **do**
  **for** $g \in R,\ a \in g$ **do**
    $M^{\ell}_{a\to g} = Emb^{\ell-1}(a)$
    $Emb^{\ell}(g) = MLP^1_j(M^{\ell}_{ne(g,1)\to g}, \ldots, M^{\ell}_{ne(g,d_j)\to g})$ *// Update rule emb. (Equations (2) and (3))*
    $M^{\ell}_{g\to a} = MLP^2_{j,i}(Emb^{\ell}(g))$
    $Emb^{\ell}(a) = \underset{(g,i):\ a=ne(g,i)}{\mathcal{A}} M^{\ell}_{g\to a}$ *// Update atom emb. (Equations (4) and (5))*

**Output layers**
**for** $a \in HB$ **do**
  $out(a) = MLP^A(Emb^L(a))$ *// Final atom prediction (Equation (6))*
**for** $g \in R$ **do**
  $out(g) = MLP^R(Emb^L(g))$ *// Final rule prediction (Equation (7))*

### 4.3. Output layers and loss functions

The output layers consist of an *atom output layer* and (possibly) a *rule output layer*.

For each atom $a$, the atom output layer takes its last available embedding, i.e. $Emb^L(a)$, and returns the prediction

$$out(a) = MLP^A(Emb^L(a)) \tag{6}$$

A similar output can be obtained for a ground rule embedding, i.e. $Emb^L(g)$, to get the ground rule prediction:

$$out(g) = MLP^R(Emb^L(g)) \tag{7}$$

The overall model is then optimized by minimizing the loss:

$$\mathcal{L} = \sum_{a\in E} L_{sup}(out(a), y_a) + \lambda \sum_{r_j\in\mathcal{T}} \sum_{g\in R^E_j} L_{sem}(out(g), r_j(\mathbf{y}_{ne(g)})) \tag{8}$$

where $\lambda \geq 0$ is a meta-parameter, $L_{sup}$ and $L_{sem}$ are standard supervised classification losses, which enforce the outputs of the model to be close to the evidence $E$. In particular, $L_{sup}$ enforces each $out(a)$ with $a \in E$ to match the available ground truth $y_a$. For instance for KGE tasks, where only positive examples are provided, the *negative softmax-loss* is used with negative examples automatically obtained through corruptions of the positive ones. On the other hand, $L_{sem}$ enforces each $out(g)$ with $g \in R^E_j = \{g : a \in E,\ \forall a \in ne(g)\}$, i.e. for the ground rules whose all neighbor atoms $ne(g)$ are in the evidence $E$, to match the evaluation of the rule $r_j$ on their overall ground truth $\mathbf{y}_{ne(g)}$. This can be done by simply evaluating the logic formula $r_j(\mathbf{y}_{ne(g)}) = r_j(y_{ne(g,1)}, \ldots, y_{ne(g,d_j)})$.

### 4.4. Discussion

*Complexity analysis.* An R2N implements a message passing schema over the factor graph obtained by grounding the logical theory. Such message passing is linear in the size of the factor graph and quadratic on the embedding size: $O(|R| \cdot d_{max} \cdot |Emb|^2)$, where $|Emb|$ is the atom embedding size, $|R|$ is the total number of ground rule nodes and $d_{max} = \max_{g\in R} |ne(g)|$ is the maximum number of atoms in the rules. If we can consider $d_{max} \cdot |Emb|^2$ a fixed constant, the computational cost is mainly determined by the $|R|$ value. As explained in Section 2.1, the number of ground rules is a polynomial depending on the number of constants in each rule $|R| = \sum_{r_j} |\mathcal{C}|^{|X_j|}$, where $|X_j|$ is the number of variables in $r_j$. The resulting complexity is in between KGE inference, which has a complexity quadratic in the number of constants (and linear or quadratic in the embedding size depending on the selected KGE), and other statistical neural-symbolic methods, which are exponential in the number of constants. However, one could rely on the sparsity of the input graph to selectively ground the logical theory around the query atoms and largely improve the scalability. We leave this study to future work to focus on the full exact grounding in this paper.

*The role of logic theories.* The employed logic theory $\mathcal{T} = \{r_1(X_1), \ldots, r_m(X_m)\}$ has the fundamental role of determining the structure of an R2N. An R2N fully grounds the rules $r_i$ on $\mathcal{C}$, without assuming that a rule holds true for a grounding, as the satisfaction or the dissatisfaction of a rule on a grounding is determined according to the available evidence. Therefore, the model learns the contexts where a rule is satisfied, driven by the ground truth. A rule that is always satisfied by the training data it will be enforced like if it was universally quantified. On the other hand, a partially violated rule will be enforced only in the contexts where it is expected to hold true. This property makes R2Ns very versatile and suitable to deal with either hard or soft rules whose percentage of fulfillment is not known in advance.

*Implicit logic rules.* Let us suppose that we know a group of atoms to be semantically connected, but without having explicitly available the logic expression of the rule correlating them, that we refer as an *implicit* rule. For instance, let $X_i$ be the set of variables for an implicit rule $r_i(P_1(\bar{x}_1), \ldots, P_l(\bar{x}_l))$, where each $\bar{x}_k \subseteq X_i$ denotes the set of variables $P_k$ depends on, for all $k \in \{1, \ldots, l\}$. The grounding process and the construction of the corresponding logic-based factor graph for $r_i$ (as described in Section 2.3) can take place similarly to fully specified FOL formulas. However, since the semantics of the rule is not specified (the logical connectives are unknown), it is not possible to determine the truth-values of the evidence groundings of $r_i$, hence the loss components given by Eq. (8) cannot be considered, e.g. $\lambda = 0$. In the experimental results, we show an use case of implicit logic rules used by R2N in Section 5.3. Even if beyond the scope of this paper, we notice that it is possible to use explainability methods to explicitly extract the logic rules a posteriori [27].

## 5. Experiments

The experiments explore the performances of the model in different contexts: symbolic reasoning with explicit logic-knowledge (Section 5.2), symbolic reasoning with implicit logic rules (Section 5.3), symbolic and sub-symbolic integration (Section 5.4) and, finally, symbolic reasoning with mined logic-knowledge on larger knowledge graphs (Section 5.5).[1] Table 1 reports the basic statistics about the datasets. Additional details on the experimental settings can be found in the Appendix.

[1] Data and code to replicate the experiments can be downloaded from: https://github.com/diligmic/R2N_KBS2024.

**Table 1**
Basic statistics of datasets. For Cora, we report the average over the five splits.

| Dataset | #Entities | #Relations | #Facts | #Degree |
|---|---|---|---|---|
| Countries | 272 | 2 | 1158 | 4.35 |
| Nations | 14 | 55 | 1992 | 142.3 |
| Kinship | 104 | 25 | 8544 | 85.15 |
| UMLS | 135 | 46 | 5216 | 38.63 |
| Cora | 700 | 9 | 18 561 | 26.51 |
| FB15k-237 | 14 541 | 237 | 310 116 | 18.71 |
| PharmKG | 7247 | 28 | 485 787 | 104.9 |
| WN18RR | 40 943 | 11 | 93 003 | 2.3 |

**Table 2**
AUC-PR metric and average train/inference times on the 3 tasks of the Countries dataset using different KGE and neural reasoning systems, R2NS indicates the relational reasoner network proposed in this paper. A bold font indicates the best method for each task.

| Task | S1 | S2 | S3 | Time |
|---|---|---|---|---|
| ComplEx | 0.993 ± 0.00 | 0.880 ± 0.03 | 0.484 ± 0.06 | 67 s/0.4 s |
| DistMult | 0.922 ± 0.05 | 0.571 ± 0.09 | 0.554 ± 0.05 | 31 s/0.2 s |
| NTP-$\lambda$ | **1.000** ± 0.00 | 0.930 ± 0.00 | 0.773 ± 0.17 | – |
| GNTP-Attn | **1.000** ± 0.00 | 0.930 ± 0.03 | 0.913 ± 0.04 | – |
| CTP | **1.000** ± 0.00 | 0.918 ± 0.01 | 0.948 ± 0.00 | – |
| NeuralLP | **1.000** ± 0.00 | 0.751 ± 0.00 | 0.922 ± 0.00 | – |
| Minerva | **1.000** ± 0.00 | 0.924 ± 0.02 | 0.951 ± 0.01 | – |
| FGNN | 0.935 ± 0.04 | 0.823 ± 0.09 | 0.560 ± 0.06 | 71 s/0.2 s |
| DeepSoftLog | **1.000 ± 0.00** | 0.977 ± 0.01 | 0.979 ± 0.01 | – |
| R2NS | **1.000** ± 0.00 | **0.992** ± 0.00 | **0.982** ± 0.01 | 125 s/1.2 s |

**Table 3**
Ablation study for R2NS in the 3 tasks of the Countries dataset, varying the number of reasoning blocks and using ComplEx as input layer. The results report the AUC-PR metric computed as average over 10 runs.

| Task | Number reasoning blocks | | | |
|---|---|---|---|---|
| | 0 | 1 | 2 | 3 |
| S1 | 0.922 ± 0.05 | 1.000 ± 0.00 | 1.000 ± 0.00 | 1.000 ± 0.00 |
| S2 | 0.880 ± 0.03 | 0.975 ± 0.01 | 0.987 ± 0.01 | 0.992 ± 0.00 |
| S3 | 0.484 ± 0.06 | 0.739 ± 0.06 | 0.848 ± 0.15 | 0.951 ± 0.03 |

### 5.1. Competitors

The proposed model is compared against state-of-the-art knowledge graph embeddings or neural-symbolic approaches. The following baselines and competitors have been used in the experiments:

- KGEs like ComplEx [28], DistMult [29], RotatE [30], HRGAT [31] and query oriented versions, like MINERVA [32].
- Neural Theorem Provers (NTP), with either greedy (GNTP-Attn, [33]) or conditional attention (CTP, [34]).
- Logic rule learning methods like NeuralLP [35], RNNLogic [36], RLogic [17], LatentLogic [37] and LPRules [38].
- Graph Neural Network (GNN) methods on factor graphs, like FGNN [39].
- ExpressGNN, a GNN-based variational approach to inference in MLNs [40].
- DeepSoftLog [41], an extension of ProbLog [42] exploiting soft-unification in the embedding space.

In the following, to distinguish if an R2N uses explicit logic rules ($\lambda > 0$) with also the semantic loss provided by Eq. (8) or just implicit logic rules ($\lambda = 0$), we will use the notation R2NS and R2NC, respectively.

### 5.2. Countries dataset

The Countries dataset (ODbL licence) [43] defines a set of countries, regions and sub-regions as basic entities. We used splits and setup from [44], which reports the basic statistics of the dataset and defines 3 tasks named $S1, S2, S3$, each requiring reasoning chains of increasing length. The task consists in predicting the unknown facts $LocIn(country, continent)$, stating country location within a continent, given the evidence in form of country neighborhoods and some known country/region locations. The model is provided with knowledge about the task like the hierarchical nature of the $LocIn$ (located in) relation,

$$\forall c \forall r \forall k \;\; LocIn(c,r) \wedge LocIn(r,k) \rightarrow LocIn(c,k) \tag{9}$$

and the manifold created across neighboring countries:

$$\forall c \forall c_1 \forall k NeighOf(c,c_1) \wedge LocIn(c,k) \rightarrow LocIn(c_1,k) \tag{10}$$

The task S1 can be solved exactly by applying the first rule. In task *S2*, some facts supporting the first rule are removed, and one has to rely on a soft application of the second rule. In task *S3*, facts supporting directly the second rule are removed, making multiple recursive applications of the second rule necessary. Table 2 reports the area under the precision–recall curve (AUC-PR) metric and one standard error for the different reasoning tasks as an average over 5 different runs. In order to ensure that the best possible results are obtained for the baseline methods, the values are taken from the original papers whenever available, for example theorem provers results have been extracted from Minervini et al. [34], while DeepSoftLog from Maene et al. [41]. R2NS shows better performance both w.r.t. plain KGEs and state-of-the-art neural theorem provers. The advantage w.r.t. KGE approaches is due to their lack of multi-hop reasoning. In fact, multi-hop reasoning is more important in S2 and S3 tasks where KGE methods clearly underperform. The advantage w.r.t. NTP is due to the fact that the grounding process in R2N is symbolic. In fact, we instantiate the same ground rules that a standard forward chaining solver will instantiate. On the contrary, both NTP and DeepSoftLog use soft-unification, which introduces more ground rules than those required by the symbolic solver (i.e. backward chaining in their case). Such ground rules are a source of noise, which becomes worse when multiple hops of reasoning are required. This is also signaled by the fact that extensions with sparser grounding techniques (e.g. CTP) tend to perform better when longer reasoning paths are required.

*Ablation study.* This experiment studies how the results on the Countries dataset are affected by the number of stacked reasoners. Table 3 reports the detailed results as average over 10 different runs. Since the S1 task is constructed to be solved by the simple application of Eq. (9), one reasoner block is enough to exactly solve the task. Moving to the more difficult S2 task, allowing more complex inference paths helps generalization. Finally, in the harder S3 task, which can be solved only via a longer reasoning chain formed by the recursive application of Eq. (10), the results are significantly improved by adding more reasoning layers. Even if not reported in the table, we observed a saturation effect when moving beyond 3 reasoning layers on this task.

### 5.3. KGE datasets: Nations, Kinship, UMLS

The Nations, Kinship, and UMLS datasets [45] (CC0 licence) are popular datasets for relational reasoning, where a set of triples (entity, relation, entity) expresses known true facts and the goal is to infer the unknown true facts. We used the setup and splits defined by [34]. In particular, the Kinship dataset is based on the kinship relationships recorded among the members of the Central Australia Alyawarra tribe, and the Unified Medical Language System (UMLS) dataset defines a set of entities representing biomedical concepts and relations. These tasks can be addressed using KGE approaches, which learn the atom representations based on the entities and relation correlations over the true facts. The main limitation of these approaches is that they fail to represent higher-order correlations among the predicates like taxonomic relations or among the constants like in transitive relations. No logic knowledge is explicitly provided for these datasets, therefore an R2NC model is instructed to correlate all predicates over each pair of constants: $r_1(P_1(x,y), P_2(x,y), \ldots, P_n(x,y))$. The model exploits the correlations to correct and improve the KGE predictions via an implicit and latent reasoning process.

**Table 4**
Results and training/inference times for KGE datasets. Missing values indicate values not reported in the original papers. The best model for each metric is shown in bold.

| Dataset | Metric | ComplEx | NTP | GNTP | CTP | Neural LP | Minerva | RNN logic | LPRules | R2NC |
|---|---|---|---|---|---|---|---|---|---|---|
| Nations | MRR | 0.749 | 0.61 | 0.658 | 0.709 | – | – | – | – | **0.911** |
| | Hits@1 | 0.627 | 0.45 | 0.493 | 0.562 | – | – | – | – | **0.871** |
| | Hits@3 | 0.858 | 0.73 | 0.781 | 0.813 | – | – | – | – | **0.930** |
| | Hits@10 | 0.998 | 0.87 | 0.985 | 0.995 | – | – | – | – | 0.993 |
| | Time | 63 s/0.02 s | – | – | – | – | – | – | – | 170 s/0.03 s |
| Kinship | MRR | 0.745 | 0.35 | 0.658 | 0.709 | 0.619 | 0.720 | 0.687 | 0.746 | **0.903** |
| | Hits@1 | 0.623 | 0.24 | 0.586 | 0.646 | 0.475 | 0.605 | 0.566 | 0.639 | **0.861** |
| | Hits@3 | 0.843 | 0.37 | 0.815 | 0.859 | 0.707 | 0.812 | – | – | **0.933** |
| | Hits@10 | 0.965 | 0.57 | 0.959 | 0.958 | 0.912 | 0.924 | 0.929 | 0.959 | **0.976** |
| | Time | 827 s/0.02 s | – | – | – | – | – | – | – | 2442 s/0.04 s |
| UMLS | MRR | 0.923 | 0.80 | 0.857 | 0.852 | 0.778 | 0.825 | 0.748 | 0.869 | **0.993** |
| | Hits@1 | 0.877 | 0.70 | 0.761 | 0.752 | 0.643 | 0.728 | 0.618 | 0.812 | **0.991** |
| | Hits@3 | 0.987 | 0.88 | 0.947 | 0.947 | 0.869 | 0.900 | – | – | **0.994** |
| | Hits@10 | **0.998** | 0.95 | 0.983 | 0.984 | 0.962 | 0.968 | 0.928 | 0.97 | 0.996 |
| | Time | 109 s/0.02 s | – | – | – | – | – | – | – | 877 s/0.12 s |

**Table 5**
AUC-PR on test for the 5 folds and global average, training/inference mean running time for the Cora dataset for different tested models. A dash indicates a missing result not reported in the original paper. Statistically significant (95%) best results reported in bold.

| Split | ExpressGNN | MLP | R2NS | R2NC |
|---|---|---|---|---|
| S1 | 0.62 | $0.830 \pm 0.007$ | **0.898** $\pm$ 0.021 | $0.823 \pm 0.015$ |
| S2 | 0.79 | $0.781 \pm 0.007$ | **0.920** $\pm$ 0.007 | $0.819 \pm 0.023$ |
| S3 | 0.46 | $0.843 \pm 0.009$ | **0.957** $\pm$ 0.003 | $0.888 \pm 0.015$ |
| S4 | 0.57 | $0.774 \pm 0.011$ | **0.923** $\pm$ 0.011 | $0.837 \pm 0.039$ |
| S5 | 0.75 | $0.838 \pm 0.006$ | **0.913** $\pm$ 0.010 | $0.800 \pm 0.019$ |
| Avg | 0.64 | $0.812 \pm 0.007$ | **0.922** $\pm$ 0.012 | $0.833 \pm 0.024$ |
| Time | – | 120.2 s/0.8 s | 1142 s/3.6 s | 809.8 s/3.6 s |

**Table 6**
Logic knowledge used by the presented models for the Cora dataset. $SamePaper(x, y)$ is defined on the union of *Author, Title* and *Venue* domains and expresses that two entities $x, y$ are related to a common paper in the known facts.

| |
|---|
| $SamePaper(x, y) \rightarrow SameAuthor(x, y)$ |
| $SamePaper(x, y) \rightarrow SameTitle(x, y)$ |
| $SamePaper(x, y) \rightarrow SameVenue(x, y)$ |
| $SameVenue(x, y) \rightarrow SameVenue(y, x)$ |
| $SameAuthor(x, y) \rightarrow SameAuthor(y, x)$ |
| $SameTitle(x, y) \rightarrow SameTitle(y, x)$ |
| $SamePaper(x, y) \wedge SamePaper(y, z) \rightarrow SameAuthor(x, z)$ |
| $SamePaper(x, y) \wedge SamePaper(y, z) \rightarrow SameTitle(x, z)$ |
| $SamePaper(x, y) \wedge SamePaper(y, z) \rightarrow SameVenue(x, z)$ |
| $SameAuthor(x, y) \wedge SameAuthor(y, z) \rightarrow SameAuthor(x, z)$ |
| $SameTitle(x, y) \wedge SameTitle(y, z) \rightarrow SameTitle(x, z)$ |
| $SameVenue(x, y) \wedge SameVenue(y, z) \rightarrow SameVenue(x, z)$ |

Table 4 reports training/inference times and the standard metrics for these tasks, where the proposed methodology outperforms KGEs and state-of-the-art reasoning systems on all datasets. The model co-trains a KGE layer, therefore adding some time overhead. In spite of the extra reasoning, running times were in the same order of magnitude for all experiments. In order to ensure that the best possible results are obtained for the baseline methods, the values are taken from the original papers whenever available, for example NTP/GNTP results have been extracted from Minervini et al. [34], while RNN Logic/LPRules have been taken from Dash et al. [38]. R2N outperforms the competitors, due to its ability to perform a more complex higher level reasoning. In particular, when explicit symbolic knowledge is not available, like in these Knowledge Graph Completion (KGC) tasks, the correlations exploited by R2NC should not be restricted to definite clauses templates exploited in backward chaining theorem provers.

### *5.4. Symbolic and sub-symbolic integration: Cora*

The Cora dataset [46] (CC0 license) defines a de-duplication task which aims at reconciling small differences in paper citations. Each paper is associated to its author, title and venue attributes. We used the setup defined by [40], which measures the performance of the model in terms of the prediction accuracy of the predicates $SameAuthor(author, author)$, $SameTitle(title, title)$, $SameVenue(venue, venue)$, detecting whether two entries refer to the same author, title and venue, respectively. Neural-symbolic methods, which focus only on the logical representation like ExpressGNN, map the textual information contained in titles/authors/venues to one binary feature modeling the presence of each term in the corresponding text. A main limitation of this class of models is that they cannot easily capture complex term correlations, required to construct powerful text similarity distances. Pure sub-symbolic models, like neural networks, can learn arbitrary complex functions to approximate the $SameAuthor, SameTitle, SameVenue$ predicates. However, they cannot easily model the relational dependencies among tasks and/or groundings.

The proposed model reasons over the latent representations generated by the underlying classifiers processing the textual information. Hence, they can simultaneously optimize the classifiers learning the text similarity predicates and the inference process defined by the available logic knowledge. Table 6 shows the logic knowledge used for this task, which is equivalent to the one used by ExpressGNN.

Table 5 reports the average AUC-PR scores and 95% confidence error over 10 different runs for the five folds. The $Avg$ split is the average of all the runs on all the folds. R2NS outperforms all the other baselines. ExpressGNN is separated by a large gap, which provides further evidence on the importance of exploiting low-level embedding functions to correlate the terms in each title, venue or author name. This is evident by also looking at the good performances of a simple MLP with no relational information. R2NS outperforms also the R2NC model, which performs better than the MLP classifier. This is a confirmation of what discussed on the advantages of using more specific knowledge, when available. The accuracy of the factor predictions is $0.822 \pm 0.031$ (95% confidence error) for factors of grounded formulas with at least one test atom, which means that the predictions of the network are consistent with the semantics of the provided formulas.

### *5.5. Large knowledge graphs: FB15k-237, WN18RR, and PharmKG*

This section studies the application of R2N to two larger scale KGs for link prediction. The first experiment is on the well known FB15k-237 dataset [47], which was constructed as a subset of Freebase, including 14505 entities and 237 relations. The second dataset is WN18RR [19], a link prediction dataset created from WN18, which is a subset of WordNet, including around 40k entities, 11 relations, 93k facts. The third considered dataset is PharmKG [31], a recently proposed biomedical Knowledge Graph, composed of around 500K individual interconnections between genes, drugs, and diseases, with a vocabulary of 28 relations and 7247 entities. As these datasets are too big for handcrafting a significant set of rules, we rely on an external miner to extract knowledge from the training data. In particular, the *Association rule Mining under Incomplete Evidence in ontological knowledge bases* (AMIE) [48] algorithm was applied to each dataset to extract a set of 100 definite clauses (top scoring clauses according to AMIE standard confidence), similarly to what done by Guo et al. [14]. The large scale of these datasets makes unfeasible to ground all the variables and formulas. Therefore, in our experiments we considered only the subset of ground formulas for which all the atoms in the body of a ground clause are true in the training data, like similarly done by other Neural-Symbolic methodologies [16,40,49].

**Table 7**
MRR, Hits@N metrics and training time on the test set of the FB15k-237 dataset.

| | MRR | Hits@1 | Hits@3 | Hits@10 | Time (s) |
|---|---|---|---|---|---|
| DistMult | 0.241 | 0.155 | 0.263 | 0.419 | 2912 |
| ComplEx | 0.259 | 0.163 | 0.292 | 0.452 | 3360 |
| RNNLogic with emb. | 0.344 | 0.252 | 0.380 | 0.530 | – |
| RLogic | 0.310 | 0.203 | – | 0.501 | – |
| LPRules | 0.255 | 0.17 | – | 0.402 | – |
| LatentLogic | 0.320 | 0.212 | 0.329 | 0.514 | – |
| R2NC | **0.347** | **0.254** | **0.382** | **0.531** | 12 960 |

**Table 8**
MRR and Hits@N metrics on the test set of the WN18RR dataset. The competitor results have been taken from Cheng et al. [17].

| | Model | MRR | Hits@1 | Hits@10 |
|---|---|---|---|---|
| KGE | TransE | 0.23 | 0.02 | 0.524 |
| | DistMult | 0.42 | 0.382 | 0.507 |
| | ConvE | 0.43 | 0.401 | 0.525 |
| | ComplEx | 0.44 | 0.410 | 0.512 |
| | RotatE | 0.47 | 0.429 | 0.557 |
| Logic based | Neural-LP | 0.38 | 0.368 | 0.408 |
| | NLIL | 0.30 | 0.201 | 0.335 |
| | DRUM | 0.38 | 0.369 | 0.410 |
| | AMIE | 0.36 | 0.391 | 0.485 |
| | RNNLogic with emb. | 0.48 | 0.446 | 0.558 |
| | RLogic | 0.47 | 0.443 | 0.537 |
| | LPRules | 0.46 | 0.422 | 0.532 |
| | LatentLogic | 0.48 | **0.497** | 0.553 |
| | R2NC | **0.49** | 0.447 | **0.559** |

**Table 9**
MRR and Hits@N results on the test set of the PharmKG dataset.

| | MRR | Hits@1 | Hits@3 | Hits@10 |
|---|---|---|---|---|
| DistMult | 0.063 | 0.024 | 0.058 | 0.133 |
| ComplEx | 0.107 | 0.046 | 0.110 | 0.225 |
| HRGAT | 0.154 | 0.075 | 0.172 | 0.315 |
| R2NC | 0.215 | 0.145 | 0.234 | 0.342 |

Tables 7–9 report the MRR and Hits@N metrics on the test set for the FB15k-237, WN18RR, and PharmKG datasets, respectively. For the FB15k-237 dataset, the baseline KGE results have been taken from Dettmers et al. [50], while the other results have been taken from the original papers, i.e. RNNLogic [36], RLogic [17], LPRules [38], and LatentLogic [37], respectively. The input layer of the R2Ns have been a RotatE and DistMult layer for FB15k-237 and PharmKG, respectively. On PharmKG, R2Ns provide a new state-of-the-art, beating the previous best model HRGAT [31] with a large margin. In any case, R2Ns significantly improve the results over the baseline models. Given the large number of relations present in FB15k-237 compared to the available knowledge, a further boost in performances is expected by manually crafting or automatically extracting a larger knowledge base to get extra coverage of the tail relations. However, this is beyond the scope of this paper.

Table 7 also reports the training time for R2N and the baseline methods measured running the models on Tensorflow2 in CPU, on a machine with 8-core Intel i7 3.4 GHz, 128 GB RAM. As expected, the training time of R2Ns is consistently larger than the training times of the baseline KGEs. Indeed, training a R2N model for this task, also implies the training of the KGE in its input atom embedding layer, plus the training of the reasoning layers.

## 6. R2N for max-product belief propagation

This section introduces a probabilistic view of the inference performed by R2N, in order to derive some theoretical guarantees about the expressiveness of the R2N reasoning process. This probabilistic formulation is not the one directly used in the experiments, but it used to highlight the expressiveness capabilities of R2Ns. In particular, Theorem 1 proves that an R2N can be exactly used to parameterize the Max-Product Belief Propagation algorithm, which is widely used as approximate inference algorithm [22]. R2Ns are built on top of a logic-based factor graphs like the ones defined by MLNs, which has at least the expressiveness of FOL (see Proposition 4.3 of Domingos et al. [23]). Therefore, in terms of logic reasoning capabilities, R2N can at least replicate the expressiveness of an MLN, when using max-product belief propagation as inference schema, which is a commonly used approximate inference algorithm. However, considering the model from a global perspective (beyond just reasoning), the neural-symbolic integration with features and embeddings makes the model more powerful than purely symbolic methods like MLNs.

As shown in Fig. 1-(b), the underlying structure of an R2N can be regarded as a factor graph. Let we call $\mathbf{x}$ a possible truth assignment to all the ground atoms in HB. A factor graph corresponds to the factorized probability distribution:

$$p(\mathbf{x}) = \frac{1}{Z}\exp\left(\sum_{g\in R}\phi_g(\mathbf{x}_{ne(g)})\right)$$

where $\phi_g(\mathbf{x}_{ne(g)})$ is a potential function corresponding to one ground rule (factor) node correlating the neighboring atom variables $\mathbf{x}_{ne(g)} \subset \mathbf{x}$. This section shows that the Belief Propagation (BP) algorithm, commonly applied to factor graphs, can be approximated by a forward pass within the R2N model.

*Belief propagation.* The (loopy) Belief Propagation algorithm defines an efficient inference schema for Probabilistic Graphical Models, like Markov Random Fields, that is based on iterative message passing. The algorithm computes an approximation of the marginal distribution for each unobserved variable, given the observed ones. Let us consider a logic based factor graph as in Section 2.3. The max-product belief propagation algorithm can be defined as follows:

$$b(x_i) = \prod_{g:\ i\in ne(g)} m_{g\to i}(x_i)\ ,$$

where $x_i$ is a discrete random variable with $i \in$ HB an atom node, whose belief $b(x_i)$ is computed as an aggregation over each neighbor factor node $g$. The message from a factor $g$ to the atom node $i$ ($m_{g\to i}$) defines the effect of that factor on the belief for that atom:

$$m_{g\to i}(x_i) = \max_{\mathbf{x}_g:\ \mathbf{x}_i=x_i} \phi_g(\mathbf{x}_g) \prod_{j\in ne(g):\ i\neq j} b(\mathbf{x}_j)$$

where $\phi_g(\mathbf{x}_g)$ is the potential associated to the factor $g$ when instantiated with the variable values $\mathbf{x}_g$ and the max is performed over all possible assignments to the variables in the factor, for a fixed $x_i$.

Lets now consider binary potentials and variables, which are the ones used in networks representing classic logic knowledge, i.e. $\phi_g(\mathbf{x}_g) \in \{0,1\}$. Hence, we have:

$$b(x_i) = \prod_{g:\ i\in ne(g)} \max_{\mathbf{x}_g:\ \mathbf{x}_i=x_i}\left[\phi_g(\mathbf{x}_g)\cdot \prod_{j\in ne(g):\ i\neq j} b(\mathbf{x}_j)\right]$$

$$= \prod_{g:\ i\in ne(g)} \max_{\mathbf{x}_g\in\mathcal{T}_g:\ \mathbf{x}_i=x_i} \left[\prod_{j\in ne(g):\ i\neq j} b(\mathbf{x}_j)\right] \tag{11}$$

where $\mathcal{T}_g$ is the set of assignments $\mathbf{x}_g$ such that $\phi_g(\mathbf{x}_g) = 1$, i.e. that satisfy the formula associated to factor $g$. The same equation can be expressed in terms of log-beliefs, which highlights the connections with neural-based implementations:

$$\log b(x_i) = \sum_{g:\ i\in ne(g)} \max_{\mathbf{x}_g\in\mathcal{T}_g:\ \mathbf{x}_i=x_i} \left[\sum_{j\in ne(g):\ i\neq j} \log b(\mathbf{x}_j)\right] \tag{12}$$

**Example 5.** This paragraph shows an example of the computation of a belief in case of factors expressing ground formulas. To keep the notation simple, we assume here that any variable only occurs in one single factor. Given two atoms $a = p_1(a_1, a_2)$, $b = p_2(b_1, b_2)$ co-occurring in a same factor $g$, the following expressions allow to recompute the beliefs for different semantics:

- Factor for logical OR between atoms: $g = a \vee b$:

$$b(x_a) = \max_{\mathbf{x}_a\mathbf{x}_b\in\{01,10,11\}:\ \mathbf{x}_a=1} \prod_{j\in\{b\}} b(\mathbf{x}_j) = \max_{\mathbf{x}_b\in\{0,1\}} b(\mathbf{x}_b)$$

$$b(x_b) = \max_{\mathbf{x}_a\mathbf{x}_b\in\{01,10,11\}:\ \mathbf{x}_b=1} \prod_{j\in\{a\}} b(\mathbf{x}_j) = \max_{\mathbf{x}_a\in\{0,1\}} b(\mathbf{x}_a)$$

- Factor for logical AND among atoms $a \wedge b$:

$$b(x_a) = \max_{\mathbf{x}_a\mathbf{x}_b\in\{11\}:\ \mathbf{x}_a=1} \prod_{j\in\{b\}} b(\mathbf{x}_j) = b(\mathbf{x}_b = 1)$$

$$b(x_b) = \max_{\mathbf{x}_a\mathbf{x}_b\in\{11\}} \prod_{j\in\{a\}} b(\mathbf{x}_j) = b(\mathbf{x}_a = 1)$$

As shown by the previous examples, the belief of a variable appearing in a single factor can be computed as function of the beliefs of the input nodes to the factor. Even if the above examples consider only simple basic formulas, the beliefs of any logic formula can be considered by following the same pattern, just considering a different set of admissible assignments. Therefore, the structure of the function is dictated by the semantic of the ground logic formula represented by the factor.

*BP and R2N.* The BP computation can be factorized in terms of the R2N modules as follows:

$$\log b(x_i) = \sum_{g:\ i\in ne(g)} \max_{\mathbf{x}_g\in\mathcal{T}_g:\ \mathbf{x}_i=x_i} \left[\sum_{j\in ne(g):\ i\neq j} \log b(\mathbf{x}_j)\right] =$$

$$= \underbrace{\sum_{g:\ i\in ne(g)} \underbrace{MLP^2_{\varphi(g),i}\Big( \underbrace{MLP^1_{\varphi(g)}\left(log_b(\mathbf{x}_g)\right)}_{\text{node to factor Eqs. (2) to (3)}} \Big)}_{\text{factor to node Eq. (4)}}}_{\text{aggregation Eq. (5)}}$$

where $\varphi(g)$ is the index of the formula for which $g$ has been instantiated, i.e. $g$ is a grounding of $r_{\varphi(g)} \in \mathcal{T}$ and $log_b(\mathbf{x}_g) = [\log b(\mathbf{x}_1), \dots, \log b(\mathbf{x}_{|ne(g)|})]$ indicates the log beliefs of neighbors of $g$ factor. In the context of an R2N, each variable $x_i$ corresponds to one ground atom e.g. $F(a, b)$. This iterative inference schema is initialized within an R2N with the beliefs computed by the input layer, which provides a solid initial guess of the values.

The following theorem shows how this factorization allows us to approximate BP with arbitrary precision, provided sufficient computational power.

**Theorem 1.** *An R2N reasoning block can exactly compute one iteration of the max-product belief propagation algorithm as a forward step within the architecture.*

**Proof.** The proof is in the Appendix. □

*Discussion.* When belief propagation can find an optimal solution, an R2N can perfectly reconstruct the same optimal output. Please note that loopy BP is often initialized with random or constant initial beliefs and it is not guaranteed to converge to a satisfactory solution. However, thanks to the neural-symbolic integration of R2Ns, the atom beliefs are instantiated by the input layer, like a KGE, which can often provide an accurate initialization of the beliefs, which are later corrected and improved by the reasoning process. For example, as described earlier in this paper, the output of a KGE is a scoring function $f_{KGE}(x_a, \theta)$, where the parameters $\theta$ and a hidden latent representation $x_a$ of the atom are used to expresses the confidence that an atom is true. The quantities $\sigma(f_{KGE}(x_a, \theta))$ and $1 - \sigma(f_{KGE}(x_a, \theta))$ can be interpreted as the initial beliefs for an atom $a$, as shown by Nickel at al. [51], where the KGE assignments are used to define a Bernoulli probability distribution. In particular, the probability of an assignment $\mathbf{y}$ with $y_a$ indicating the assignment to atom $a$ can be computed as:

$$p(\mathbf{y}|\mathbf{x}, \theta) = \prod_a p(y_a|x_a, \theta) = \prod_a \begin{cases} \sigma(f_{KGE}(x_a, \theta)) & if\ \ y_a = 1 \\ 1 - \sigma(f_{KGE}(x_a, \theta)) & if\ \ y_a = 0 \end{cases}$$
$$= \prod_a \left[1 - y_a + \sigma(f_{KGE}(x_a, \theta))(2y_a - 1)\right]$$

In practice, an R2N usually employs a larger atom embedding space than the one required to represent only the beliefs of a binary variable. As a consequence, the model can perform complex atom transformations and can learn how each formula should be accounted depending on the input constants and predicates. In this scenario, the internal representation $x_a$ of the atom of a KGE can be directly passed to the reasoning steps, without the information bottleneck of compressing it to scalar beliefs.

## 7. Related work

Relational Reasoning Networks bridge ideas from different research areas, like logic, neural networks, probability, and, as a result, they present several connections with different AI models. In the following, the links between R2Ns and prominent work in different areas are highlighted.

*R2Ns and (probabilistic) reasoning.* The capability of make inference is a fundamental property of AI systems whose behavior could be considered intelligent, and possible ways on how to automatize the process of reasoning have received a lot of attention over the years [52]. In forward chaining (or forward reasoning) novel true facts are produced as a result of the application of one or more inference steps, starting with a set of known true facts. In the same spirit, an R2N performs one or more sub-symbolic reasoning steps by manipulating a set of input fact embeddings into a set of refined new ones, accounting for the relational structure of the task. Probabilistic reasoners define a more flexible inference process than their counterparts based on standard logic. For example, Statistical Relational Learning approaches like Markov Logic Networks (MLN) [23] translate a FOL theory into an undirected graphical model. Probabilistic Soft Logic [53] tried to overtake the scalability limitations of MLNs by defining a tractable fraction of FOL and relaxing inference using fuzzy logic. R2Ns exploit the same graph construction of MLNs but inference is performed faster as a forward step of a neural architecture.

*R2Ns and neural-symbolic AI.* Neural-symbolic methods [6] bridge the reasoning capabilities of logic reasoners with the ability of sub-symbolic systems to deal with the feature-based representations that typically represent sensorial inputs like video, images or sounds. In particular, neural-symbolic distillation methods like Semantic-based Regularization [54], teacher–student setups [55] and Logic Tensor Networks [56] inject logic knowledge into the network weights by enforcing the knowledge on the predictor outputs. In spite of their simplicity, these methods are powerful when the logic knowledge should be applied to the entire output space in a flat fashion, unlike what is needed to obtain a full probabilistic reasoning process. R2Ns define a more flexible employment of the logic knowledge, since reasoning over latent representation learns how and where to apply the available knowledge. Neural Logic Machines (NLM) [57] exploit neural networks and logic programming to carry out both inductive learning and logic reasoning tasks. However, NLM could not be applied to Knowledge Graph Embeddings tasks, where the relational information is not fully known. Further NMLs directly compute output Boolean values, which strongly limits the flexibility of reasoning under uncertainty, while reasoning in R2Ns happens at latent level. Relational Concept Bottleneck Models (R-CBMs) [58] extend interpretable deep learning methods like Concept Bottleneck Models (CBM) to relational domains. This class of methods is based on the definition of a set of intermediate concepts, which forms the vocabulary used to provide explanations. R2Ns are strictly more expressive than R-CBMs since they can represent the concepts in their embeddings, but they are not limited to that. On the contrary, the mapping from embeddings to the final decision is less interpretable in R2Ns than in R-CMBs. Recently, concept-based reasoners [59,60] have explicitly linked neurosymbolic reasoning and concept-based modeling by learning logic-based decision layers built on neural concept representations. Another class of neural-symbolic approaches, like Deep ProbLog [8], Semantic Loss [61], Deep Logic Models [62] and Relational Neural Machines [63,64] define directed or undirected graphical models to jointly train predicates approximated by deep learners and reasoning layers in a single differentiable architecture. Exact inference over the graphical models is generally intractable and these systems rely on heuristics to apply to any real-world problem [65,66]. Lifted Relational Neural Networks [67] and Neural Theorem Provers (NTP) [33,34,44] realize a soft forward or backward chaining via an end-to-end gradient-based schema. NTP allows soft-unification among symbols using a tensorial representation to allow a more flexible matching. However, this introduces scalability issues as reasoning paths cannot be easily pruned without relying on heuristics. Neural LP [35] learns rules using differentiable operators defined in TensorLog [68], which is a framework to get advantage of deep learning computational power to perform probabilistic logical reasoning. Unlike this class of models, R2N approximates relational inference using a forward step in a neural architecture and can scale up to much larger problems. DeepSoftLog [41] is an extension of DeepProbLog [8], where entities and relations can now be embedded in a latent space and matched to each other using soft-unification. However, DeepSoftLog maintains the exact inference approach of DeepProbLog, making it unfeasible in large scale settings.

*R2Ns and KGEs.* With the emergence of Knowledge Graphs, representation learning has played a pivotal role in learning embedding spaces, which allow efficient inference and query answering. Knowledge Graph Embeddings, initially built from the statistical co-occurrences of entities and relations, have been later enriched with semantic information carried by higher-level logic knowledge to improve embedding accuracy and explainability. Two main tasks in the context of knowledge graphs are link prediction and complex query answering. The latter assumes to have a complex query formulation, whose structure is known a priori. For instance, approaches like Query2Box [69] and BetaE [70] use KGs for query answering by modeling multi-hop reasoning in the latent space. MINERVA [32] faces this task by walking on the KG conditioned on a query, and then by stopping when reaching the answer node. Complex query answering using logic knowledge can scale to large KGs, because the chain of inference is fixed and fully known from the data. However, this process is very different from the adaptively learned chain of reasoning of R2Ns. Indeed, the link prediction task requires to establish if pairs of entities are related, and potentially every inference chain can serve the purpose. As an example of this line of research, both probabilistic Logic Neural Networks [71] and the work by Niu at al. [72] inject logic knowledge into a KGE. However, these approaches require strong assumptions on the form of the distributions to make inference tractable. RNNLogic [16] co-learns a set of horn clauses per query with a recurrent neural network, which are used by a log-linear layer to assign a score to the query. This process corresponds to a one-hop reasoning which is fundamentally different from the multi-hop reasoning process instantiated by the R2N message passing architecture. In order to generalize the mining of purely observed rules, RLogic [17] learns schema-level logical rules via a scoring model based on predicate representation learning, which is trained from a small set of sampled rule instances. Even if more data efficient, this strategy may introduce several logic-based biases to guide the reasoning steps, hence affecting the generalization capability of the overall model. LatentLogic [37] learns the logic rules by using a pre-trained variational auto encoder in the latent space. R2N makes a very different use of the embedding space, which is used to refine the logic atoms as a sub-symbolic inference process.

*R2Ns and GNNs.* R2N is related to a recent line of research using Graph Neural Networks (GNNs) [25] to approximate inference in PGMs. Standard GNNs are examples of Message Passing Neural Networks [26], whose expressiveness can be interpreted in terms of two-variable counting logic [73], a small fragment of First-Order Logic. In order to overtake the limitations of standard GNNs in approximating inference, some authors have proposed hybrid models where GNNs are integrated to replace one step of a traditional inference algorithm such as belief propagation [74]. Other work has instead attempted at defining higher-order GNNs [75] which yield expressive models at the cost of a combinatorial explosion of the higher-order edges. R2Ns stand in the middle of standard GNNs and higher-order GNNs as the logic provides a bias on which "higher-order edges" (i.e. factors) to instantiate. The proposed model can indeed be seen as a higher-order GNN, where the layers computing the factor representations process the higher-order information. In this regard, R2N is related to Factor Graph Neural Networks (FGNN) [39], but applied to structures that represent grounded FOL logic theory. Unlike FGNNs, R2Ns are aware of the order of the in-going and outgoing edges in the factor graph. In particular, $MLP^1_j$ concatenates its in-going messages following the order in which atoms appear in a rule, and one different $MLP^2_{j,i}$ is specialized to correctly predict the embedding for the $i$th element in the list of the outgoing neighbors. Furthermore, unlike what done by FGNNs, R2N is designed to work in a neural-symbolic environment, where the input representations are also co-developed during training. R2Ns are also related to recurrent graph neural networks (such as graph LSTMs) with constrained gating mechanisms [76]. Here, activations in both the spatial and temporal domains are constrained to be coherent.

Other GNNs have been defined to process the KG topological structure. For example, ExpressGNN [40] attempts to overtake the scalability limitations of probabilistic reasoning approaches using a GNN defined over the KG to embed the constants. Since the employed GNN does not have enough expressive power to perform reasoning, as there are isomorphic substructures in the KG that cannot be distinguished, reasoning is performed via an external variational process. A similar relational structure is considered by Discriminative Gaifman models [77], which translate FOL formulas into a set of features. Unlike this class of models, the proposed architecture directly compiles the relational structures required to approximate the reasoning process into the model architecture. Therefore, R2N can embed the constants and atoms via knowledge in the forward step of the network without employing an external inference step.

## 8. Conclusions

This paper presents a neural architecture to perform relational reasoning using latent representations. The model can be applied on top of different input feeds like KGEs or the embeddings computed by deep neural networks. The presented model provides a flexible and expressive platform for neural-symbolic integration, which can be used either when logic knowledge is explicitly available or when the expression of the relational knowledge is not fully known.

While the scalability of the presented model is a huge step forward to classical Statistical Relational Learning methods, application on large domains can still be impractical when working on a fully grounded HB, as the number of possible ground atoms grows polynomially on the arity of the considered relations. An accurate approximation of the inference process on the relevant subset of ground atoms is a promising line of research, currently explored by the authors.

## CRediT authorship contribution statement

**Giuseppe Marra:** Writing – review & editing, Writing – original draft, Visualization, Validation, Supervision, Software, Resources, Project administration, Methodology, Investigation, Funding acquisition, Formal analysis, Data curation, Conceptualization. **Michelangelo Diligenti:** Writing – review & editing, Writing – original draft, Visualization, Validation, Supervision, Software, Resources, Project administration, Methodology, Investigation, Funding acquisition, Formal analysis, Data curation, Conceptualization. **Francesco Giannini:** Writing – review & editing, Writing – original draft, Visualization, Validation, Supervision, Software, Resources, Project administration, Methodology, Investigation, Funding acquisition, Formal analysis, Data curation, Conceptualization.

## Declaration of competing interest

The authors declare the following financial interests/personal relationships which may be considered as potential competing interests: Michelangelo Diligenti reports financial support was provided by EU Framework Programme for Research and Innovation Euratom.

## Acknowledgments

We thank Filippo Guerranti, Stefano Fioravanti, Caterina Graziani for their help in running one experiment presented in this paper. This project has received funding from the European Union's Horizon 2020 HumanAI-Net research and innovation program under grant agreement No 952026. This work was also supported by TAILOR, a project funded by EU Horizon 2020 research and innovation programme under GA No 952215 and by the EU Framework Program for Research and Innovation Horizon under the Grant Agreement No 101073307 (MSCA-DN LeMuR). This research has also received funding from the KU Leuven Research Fund (STG/22/021, CELSA/24/008) and from the Flemish Government under the "Onderzoeksprogramma Artificiële Intelligentie (AI) Vlaanderen" programme. Francesco Giannini has been supported by the Partnership Extended PE00000013 - "FAIR - Future Artificial Intelligence Research" - Spoke 1 "Human-centered AI".

## Appendix. R2N as belief propagation

### *Logic-based belief propagation*

**Proof of Theorem 1.** In this section, we recall Theorem 1 and provide its proof.

**Theorem 1.** *An R2N reasoning block can exactly compute one iteration of the max-product belief propagation algorithm as a forward step within the architecture.*

**Proof.** Assume without loss of generality that a two-dimensional embedding space is used to represent the atoms, such that these representations can be the log-beliefs for the $\{0,1\}$-assignments for each atom at the previous iteration. For any variable $x_i$, an R2N reasoning block can reproduce one iteration of the max-product BP algorithm if:

1. for any $g$, the factor update function ($MLP^1_{\varphi(g)}$) takes the log-beliefs of all the variables as input and computes a factor representation that collects for each assignment of $\mathbf{x}_g$ and variable $x_i \in \mathbf{x}_g$: $\sum_{j\in ne(g):\ i\neq j} \log b(\mathbf{x}_j)$ , which is a simple function of the input log-beliefs. The resulting factor representation collects these values in a vector of size $|ne(g)| \cdot 2^{|\mathcal{T}_g|-1}$.
2. The factor-to-node Multi-Layer Perceptron ($MLP^2_{\varphi(g),i}$) selects the maximum over the input values, corresponding to sum of the log-belief of the assignments $\mathbf{x}_g \in \mathcal{T}_g$, with $\mathbf{x}_i = x_i$.
3. The final belief of a node is computed as summation over a per-factor contribution in the BP algorithm.

Since an MLP with at least one hidden layer and a sufficient number of hidden neurons is a universal function approximator [78,79], it trivially follows that it is possible to train the MLPs to approximate the first two steps with arbitrary precision, if provided with sufficient computational power, and the factor representation has at least $|ne(g)| \cdot 2^{|\mathcal{T}_g|-1}$ values. Finally, if the aggregation operator in Eq. (5) is selected to be the summation over the factor contribution, the factor aggregation step of BP is perfectly replicated by an R2N. □

In addition, it is possible to provide a tighter definition of the minimum computational power of the MLP architectures, instead of the over-estimates provided by the universal approximation theorem. In particular, the $MLP^1$ has to compute different summations over its inputs, such that a single layer network with linear neuron activation already provides sufficient computation power. On the other hand, the following proposition provides a direct estimate of the number of layers and neurons needed by the $MLP^2$ networks to compute the selection of the maximum element in Eq. (12) using RELU neuron activations.

**Proposition 1.** *For arbitrary real valued feature matrix $\mathbf{X} \in \mathbb{R}^{m\times n}$ with $x_{ij}$ as its entry in the $i$th row and $j$th column, the feature mapping operation $x^\star = [\max_j x_{ij}]$ can be exactly parameterized with a $2\log_2 n$-layer neural network with RELU as activation function and at most $2n$ hidden units. See Zhen et al. [39] for a proof.*

## Experimental details

*Experimental settings.* All the employed datasets do not include personal data. The size of the embedding space has been selected by maximizing the target metric on the validation set for each experiments, using the grids reported in the provided code package.

*Countries experiment.* In the Countries experiment (Section 5.2), an R2NS model with 3 stacked reasoning layers was trained for 300 epochs, Adam optimizer with initial learning rate equal to 0.01 and the reasoning embedding size was selected via a validation set in the $[5, 50]$ range. Since FGNN requires Boolean inputs, the KGE output (i.e. the true belief for each atom) is directly fed to the FGNN, which performs the reasoning process using the same knowledge used by the R2NS.

*Knowledge graph completion experiment.* In the knowledge graph completion experiment (Section 5.3), the different KGE methods have been tested for each dataset, searching the embedding size maximizing the Mean Reciprocal Rank (MRR) metric on the validation set over a grid in the $[10, 100]$ range. The best performing KGE was employed as first layer of the model and co-trained for 700 epochs, Adam optimizer with an initial learning rate equal to 0.01 and reasoning embedding size selected using the validation set on the same grid $[10, 100]$.

*Cora experiment.* In the Cora experiment (Section 5.4), the input text was represented by a bag-of-word representation and processed by an input MLP with 2 layers and 35 neurons per layer. The classifier and reasoning layers have been co-trained for 400 epochs, using the Adam optimizer with an initial learning rate equal to 0.01, and reasoning embedding size equal to 35 (validated on the validation set). The same architecture has been used for the competitor MLP.

*PharmKG.* : the input KGE layers have been pretrained for 100 epochs.

*FB15k-237, WN18RR and PharmKG.* a single reasoning layer with residual connection has been trained for another 100 epochs with the Adam Optimizer and an initial learning rate equal to 0.01 using an embedding size equal to 200.

## Data availability

Data and code to replicate the experiments can be downloaded from: https://github.com/diligmic/R2N_KBS2024.